\documentclass[10pt,twocolumn,letterpaper]{article}

\usepackage{cvpr}              

\usepackage{graphicx}
\usepackage{amsmath}
\usepackage{amssymb}
\usepackage{booktabs}
\usepackage{hhline}
\usepackage[accsupp]{axessibility}

\usepackage[pagebackref,breaklinks,colorlinks]{hyperref}

\usepackage[capitalize]{cleveref}
\crefname{section}{Sec.}{Secs.}
\Crefname{section}{Section}{Sections}
\Crefname{table}{Table}{Tables}
\crefname{table}{Tab.}{Tabs.}

\def\cvprPaperID{13} 
\def\confName{CVPR}
\def\confYear{2023}

\begin{document}

\title{ VisiTherS: Visible-thermal infrared stereo disparity estimation of human silhouette }

\author{
Noreen Anwar, 
Philippe Duplessis-Guindon, 
Guillaume-Alexandre Bilodeau \\
LITIV, Polytechnique Montréal \\
{\tt\small {noreen.anwar@polymtl.ca, philippe.duplessis-guindon@polymtl.ca, gabilodeau}@polymtl.ca}
\and
Wassim Bouachir \\
Data Science Laboratory, University of Québec (TÉLUQ)\\
{\tt\small wassim.bouachir@teluq.ca}
}

\maketitle

\begin{abstract}
   This paper presents a novel approach for visible-thermal infrared stereoscopy, focusing on the estimation of disparities of human silhouettes. Visible-thermal infrared stereo poses several challenges, including occlusions and differently textured matching regions in both spectra.  Finding matches between two spectra with varying colors, textures, and shapes adds further complexity to the task. To address the aforementioned challenges, this paper proposes a novel approach where a high-resolution convolutional neural network is used to better capture relationships between the two spectra. To do so, a modified HRNet backbone is used for feature extraction. This HRNet backbone is capable of capturing fine details and textures as it extracts features at multiple scales, thereby enabling the utilization of both local and global information. For matching visible and thermal infrared regions, our method extracts features on each patch using two modified HRNet streams. Features from the two streams are then combined for predicting the disparities by concatenation and correlation. Results on public datasets demonstrate the effectiveness of the proposed approach by improving the results by approximately 18 percentage points on the $\leq$ 1 pixel error, highlighting its potential for improving accuracy in this task. The code of VisiTherS is available on GitHub at the following link: \url{https://github.com/philippeDG/VisiTherS}. 
\end{abstract}

\section{Introduction}
\label{sec:intro}

The objective of this paper is to propose a method for estimating pixel disparities between a visible image and a thermal infrared image. The idea of combining these two types of images is to benefit from each of them for tasks, such as object detection. If both images form a stereo pair, estimating disparity allows depth estimation that can further be used to improve detection itself or subsequent tasks, like tracking. The estimation of these pixel disparities can be used to align the visible-thermal infrared stereo thereby generating an augmented image. These augmented images are particularly useful in challenging scenarios such as low light, fog, and smoke, and can significantly enhance the accuracy of object detection and tracking. In contrast to classical stereo, pixel pattern-based approaches are insufficient in visible-thermal infrared stereo, highlighting the need for more advanced techniques for estimating pixel disparities.

\begin{figure}[t]
\centering
\subfloat[RGB rectified image]{\includegraphics[width=1.5in]{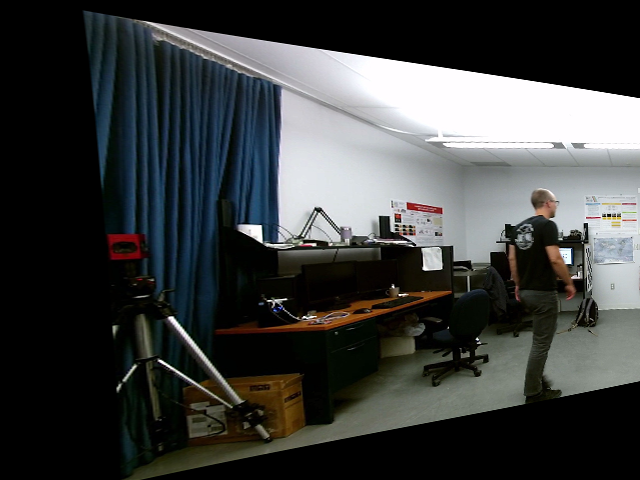}\label{rgb_original}}
\subfloat[LWIR rectified image]{\includegraphics[width=1.5in]{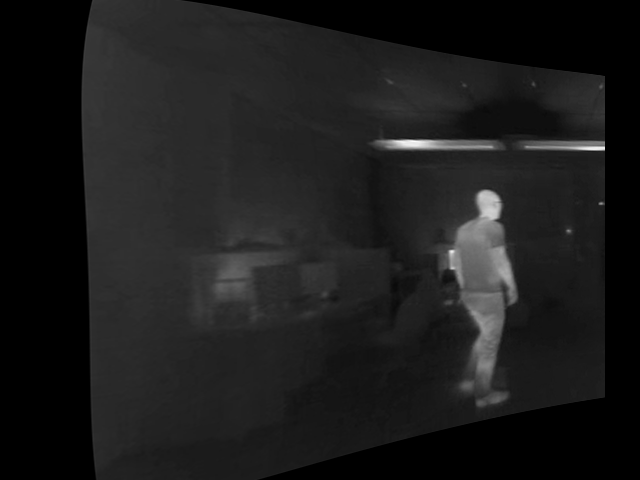}\label{lwir_original}}
\caption{A rectified visible and thermal infrard image from the LITIV 2018 dataset \cite{st2019online}.}
\label{patern_ex}
\end{figure}

To deal with this particular challenge, we focus on estimating disparities of human silhouettes. We assume that those disparities are estimated from a sparse set of points, that is, our method is designed for sparse stereoscopy.  Human silhouettes can be captured in both visible and thermal images, where the silhouette in thermal images is formed from the body's heat emission, and in the visible images by the color on the person. In this case, relying solely on pixel patterns as in classical stereo is insufficient. For example, the shirt logo present in the visible image (Figure \ref{rgb_original}) is absent in the thermal image (Figure \ref{lwir_original}). In classical stereo, the shirt logo would have been an effective mean of estimating the disparity. We can also observe that the heat emitted through the shirt is unevenly distributed across surface. Consequently, the thermal image displays some intensity variation in the shirt region, while the visible image depicts uniform black coloring. We aim to propose a precise and efficient method that can estimate the disparity between the pixels in two human silhouettes. 

This paper introduces a new convolutional neural network (CNN) architecture, called VisiTherS (standing for Visible-Thermal infrared Stereo), for estimating the disparity between visible and thermal image pairs. We propose to use a high-resolution network for extracting features as we believe that it can capture better the relationships between the pixels in both types of images. For this purpose, we selected HRNet to obtain a series of feature maps with strong semantic meaning at different scales. We investigated two ways to use the feature maps. First, we concatenated the features at the different scales of the last stage, which are adjusted to match the highest resolution feature map size. The resulting feature maps can take advantage of both high-resolution information and multi-scale information, providing a more comprehensive representation of the input data.Second,we concatenated high-resolution feature maps of two stages, thus retaining only the high-resolution information. We show that both of these strategies to exploit high-resolution features  significantly improve results compared to the best SOTA methods. 

Our proposed method consists of two HRNet streams, where each patch of the image in the stereo pair has its own feature extractor.While both streams have the same structure, but there is  no weight sharing between them. VisiTherS takes two small square patches as input and extracts features from them, resulting in a feature vector for each image patch. To enhance the robustness of our network for predicting disparities, we employ two fusion techniques on the feature vectors. Firstly, we perform a correlation product between both vectors, forwarding the result to the correlation head. Secondly, we perform a concatenation between the two vectors, forwarding the result to the concatenation head. Better results are generally obtained by using the two fusion techniques simultaneously compared to only using correlation or concatenation. The correlation and concatenation heads consist of fully connected layers outputing the probability of both patches being the same or not. Each classification head has its own loss function, and during testing, we employ both classification heads to obtain the disparity predictions.

Our contributions can be summarized as follows: 
 \begin{itemize} 
\item 
 We propose VisiTherS, a new CNN architecture based on two streams composed of a high-resolution convolutional neural network feature extractor. Our architecture extracts features from both image domains and uses two fusion processes to compute the probability of the input patches being the same.
 \item 
 Our findings show that our model is highly effective in performing disparity estimation between visible and thermal image pairs and that high-resolution features are a good choice for this kind of task. This represents a significant improvement over existing approaches and highlights the potential of our novel CNN architecture in advancing the field of disparity estimation. 
\end{itemize}


\section{Literature review}
Stereo estimation can be achieved through two primary approaches: sparse stereo and dense stereo. These two approaches are the main methods used to perform the stereo estimation.  Sparse stereo estimation involves selecting two regions from the original images, rather than inputting the entire images into the network. That is, only small patches around the disparity points are fed into the network. The objective is to find the corresponding patch in the other image. The disparity is calculated by measuring the pixel distance between the coordinates of these two patches. Since dense disparity labels are not required, this approach is applicable to both dense and sparse datasets, although it is generally slower in the case of dense stereo estimation.

Several papers have explored this approach in visible-visible stereo, including the pioneering work of Zbontar and LeCun \cite{7298767}, where a CNN is used to learn the similarity between a $9\times 9$ region on the left and right images, with the goal of determining the disparity between these regions. Luo et al. \cite{efficientDL} built upon this approach by creating a feature vector for the left image patch and a feature volume for the right patch and using correlation products to calculate the probability distribution of the disparity. Kendall et al. \cite{kendall2017endtoend} proposed the GC-Net, which was the first end-to-end architecture using a Siamese network for feature extraction and 3D convolution for disparity mapping. Other methods have then been developed, including those using spatial pyramid pooling modules for feature extraction, and hourglass networks for cost volume regularization and disparity regression \cite{chang2018pyramid}.

In dense stereo, disparities are estimated for each pixel of the images. To effectively train a machine learning model and reduce the likelihood of it overfitting, it is necessary to have datasets that contain a large number of densely labeled examples. The first that proposed an end-to-end dense stereo model were Mayer et al. \cite{mayer2016large}. This work had a huge impact on the field since they created a densely annotated dataset FlyingThings3D. This dataset consists of images having a disparity value at every pixel, which leads to a lot of subsequent work using the end-to-end method. Their method, called DispNet, is inspired by FlowNet \cite{dosovitskiy2015flownet} for compression and decomposition, respectively. The compression part is built with convolutions that result in a final reduction factor of 64. The decompression then resizes the disparity maps gradually in a non-linear way, taking into consideration the characteristics in the compression step. The final result of the network is a disparity map with the same image size as the input image.


Prior to the rise of neural networks, Visible-thermal infrared stereo relied on matching feature points, often using SIFT\cite{Lowe2004} as a feature descriptor. MSIFT\cite{5995637} was then introduced to improve the correlation between RGB channels in RGB (visible)-NIR (Near-infrared) pairs of images. However, some methods have opted to use window-based methods, such as mutual information \cite{466930}, HOG \cite{HOG}, SSD \cite{bilodeau2014thermal}, LSS \cite{torabi2011local}, to find image matches. Among these window-based methods, Bilodeau et al. \cite{bilodeau2014thermal} found that mutual information is the most accurate approach \cite{st2019online}.

In recent studies of visible-thermal infrared stereo, Beaupré et al. \cite{Beaupre_2019_CVPR_Workshops} proposed a novel method using two Siamese networks to compute the disparity from visible to thermal and vice versa. The Siamese networks have shared parameters, and their architecture is similar that of Luo et al\cite{efficientDL}. The method involves comparing a small patch of a visible image with a patch in the thermal image of the same height, but of the full width of the original image. The correlation is done with every possible translation to find the corresponding disparity. The same principle applies to the other Siamese network, however the small patch correspond to is the thermal image at the given disparity while the wider image is the visible image. To select the final disparity, a summation layer is used, to sum up, the prediction vector from each network branch, with the final disparity being the maximum element.

In a subsequent work by Beaupré et al. \cite{beaupre2021domain}, a modified approach was proposed,  yielding to significant improvements over the previous method. Unlike the previous method, this approach does not share weights between the two feature extraction branches. This change was made due to the dissimilar nature of the two types of images used in visible-thermal infrared stereo matching. Unlike the typical inputs used in Siamese networks, the visible and thermal images are dissimilar in terms of color, shape, and contrast. The only aspect they have in common is the shape of the objects, which is not even exactly the same due to the differences in how the images are captured. Therefore, parameter sharing between feature extractors is not appropriate in this case. This approach served as an inspiration for our work. In the work of Duplessis-Guindon et al. \cite{duplessis20224d}, an approach was proposed for estimating the disparity of people in a scene using segmentation masks obtained from both visible and thermal images. Masks helped estimate the disparities at the object boundaries.

Visible-infrared stereo matching is not limited to thermal infrared, as there have been studies on Visible-Near infrared (NIR) stereo as well. Aguilera et al. \cite{Aguilera2016} investigated the effectiveness of three different CNN architectures compared to the traditional methods mentioned earlier for this task. Building on their previous work, Aguilera et al. \cite{Aguilera2017} introduced quadruplet networks that take two matching pairs of images, providing two pairs of positive examples and four pairs of negative examples for training. However, similarly to visible-thermal infrared stereo, there is a shortage of datasets for Visible-NIR stereo. To address this problem, Zhi et al. \cite{Zhi_2018_CVPR} created a method that transforms a visible image into the NIR spectrum and uses the resulting image for self-supervised learning.

\begin{figure*}[h]
\begin{center}
\centerline{\includegraphics[width=5in]{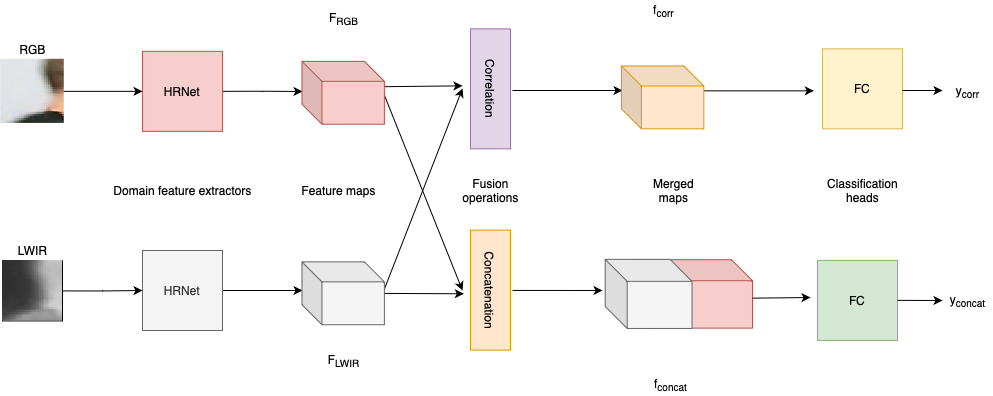}}
\caption{Overview of our proposed network architecture. RGB: visible, LWIR: Thermal infrared. Features are first extracted by two independent streams for RGB and LWIR. Features are then combined together using concatenation and correlation. Correlated features and concatenated features are then fed to two separate classification heads. Classification results are then combined (not shown in the figure).}
\label{sCNNs}
\end{center}
\end{figure*}

\section{Proposed method}

 Our method is inspired by the work of \cite{beaupre2021domain}. Figure \ref{sCNNs} visually depicts the overall architecture of our model. It is composed of two streams, one for the visible (RGB) and one for thermal infrared (LWIR) patches. In both, features are extracted using a high-resolution CNN. Features are then fused and patches are classified. Our architecture is detailed in the following.
          
\subsection{Feature extractor}

In this section, we explain in detail the feature extraction part of our architecture. It requires two patches as input, an RGB and an LWIR patch. These patches are sized $36\times 36$ to capture the surrounding context of the image around a point where we wish to calculate disparity. These patches are referred to in the following as $P_{RGB}$ and $P_{LWIR}$. As shown in Figure \ref{sCNNs}, each patch is processed by its own feature extractor with different learned weights. Each feature extractor outputs a $36 \times 36 \times 64$ feature map represented by $F_{RGB}$ and $F_{LWIR}$ as illustrated in Figure \ref{sCNNs}.

For feature extraction, we selected HRNet\cite{yuan2021hrformer} to obtain high-resolution features. This is motivated by the fact that visible and thermal infrared are different, and we believe that more expressive feature maps are required to match them. Traditional CNN backbone architectures reduce resolution between convolution layers, leading to less information in the final feature maps. HRNet major objective is to align input and output resolution. HRNet maintains resolution after each convolution and each stage adds a new feature map scale. The network output is a concatenation of these feature maps. All feature maps are resized to match the original input size. The final feature map, therefore, has a large number of channels. The original HRNet\cite{sun2019deep} network performs a series of convolutions on this final feature map to reduce its dimensions. However, our goal in introducing this feature extractor is to have the best possible resolution. We therefore only scaled the number of channels to have as output a feature map of size $36 \times 36 \times 64$. The last dimension of the feature map represents the number of feature channels. In our HRNet architecture, we removed the upper layers from the original and kept only the first three stages. 

\begin{figure*}[!t]
\centering
\subfloat[Concatenation of feature maps of various scales from the last stage. ]{\includegraphics[width=6in]{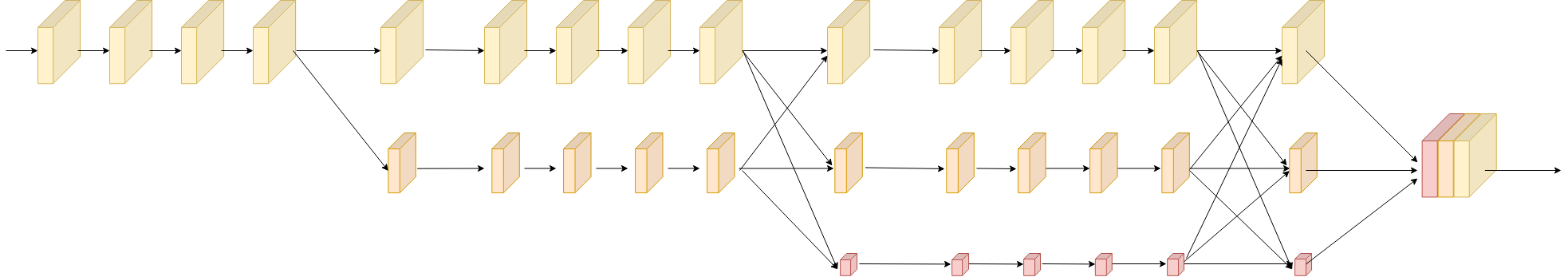}\label{fig:a}}

\subfloat[Concatenantion of feature maps from the last two stages.]{\includegraphics[width=6in]{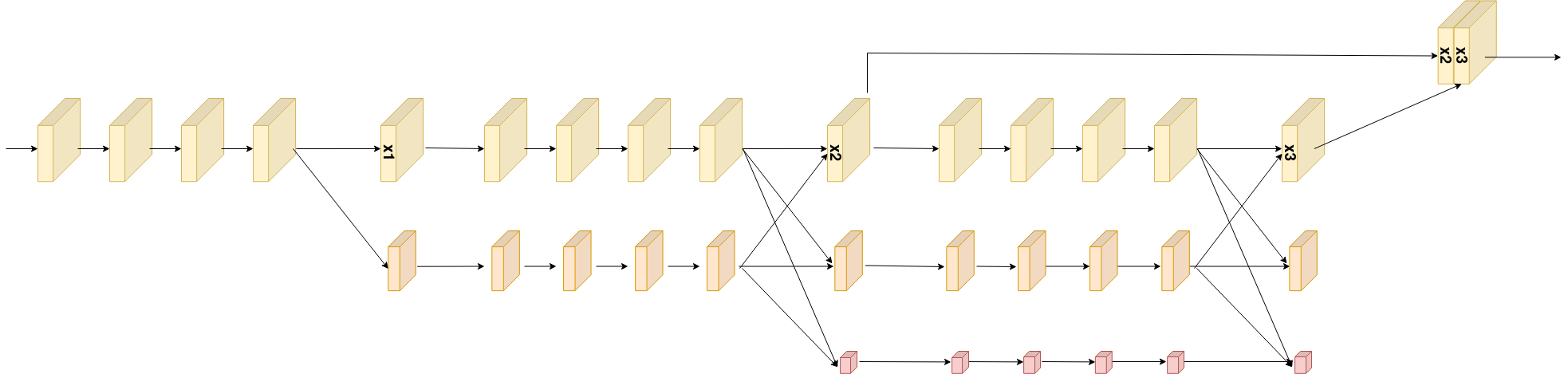}\label{fig:b}}
\caption{Our proposed versions of feature extractors. For $36 \times 36 $ image patches, neglecting the number of channels, yellow feature maps are $36 \times 36 $, orange feature maps are $18 \times 18$,  and red feature maps are $9 \times 9 $.}
\label{both}

\end{figure*}

We investigated two ways of exploiting the feature maps generated by HRNet. In the first, we concatenate features from several scales at the last stage. This is illustrated in Figure \ref{fig:a}. In the figure, yellow feature maps are $36 \times 36 $, orange feature maps are $18 \times 18$, and red feature maps are $9 \times 9 $. To obtain the concatenated feature map, we concatenated the three feature maps of the last stage and adjust their sizes to match the highest resolution. This results in a concatenated output of the three multiscale feature maps, which formed the final feature map $F_{RGB}$ or $F_{LWIR}$, depending on the stream. 

In the second way presented in Figure \ref{fig:b},  we concatenate the highest resolution features from several stages. More precisely, we are concatenating the high-resolution feature map of the last stage with that of the previous stage. Therefore, we only keep high-resolution information. This results in a concatenated output of the two high-resolution feature maps, which formed the final feature map $F_{RGB}$ or $F_{LWIR}$, depending on the stream.

\subsection{Classification heads}
In our proposed method, we employ two distinct fusion operations on feature maps, namely correlation and concatenation, as described in \cite{guo2019group}. These fusion operations are widely used in disparity estimation for integrating image features. While both operations have their advantages, each also presents certain limitations. Specifically, the correlation fusion operation is characterized by its computational speed and memory efficiency; however, it may result in the loss of some features from both spectra during the fusion process. On the other hand, the concatenation operation does not lead to any loss of features, but it entails a trade-off between the computational time and memory space required for its implementation. The correlation operation outputs a  $36 \times 36 \times 64$  feature map, represented by $f_{corr}$ in Figure \ref{sCNNs}. The concatenation operation outputs a $72 \times 36 \times 64$  feature vector, which is represented by $f_{concat}$ in Figure \ref{sCNNs}. Both $f_{corr}$ and $f_{concat}$ are going through separate fully connected networks (FCNs). The weights are not shared between each fully connected network and each output a classification vector. These are represented by $y_{corr}$ and $y_{concat}$ in Figure \ref{sCNNs}. Both FCNs generate a 2D probability vector and this vector represents the likelihood that two patches are either identical or different.

\subsection{Training losses}

The network can learn by training on two corresponding image patches of $36\times 36$ pixels ($P_{RGB}$ and $P_{LWIR}$), one for the visible spectrum and one for the thermal spectrum. During inference, the network is fed with a $36\times 36$ patch in the visible spectrum and it tries to locate the corresponding patch within a larger thermal image patch. In other words, the network learns to associate the two types of images and can use this knowledge to identify the location of a visible patch within a thermal patch. 

To train our network, we employ two separate loss functions, one for the correlation head and another for the concatenation head. This allows us to optimize the network performance based on both fusion schemes. They are given by
\begin{equation}
    loss_{corr} = -1/N \sum_{i=1}^N gt^i log(y_{corr}^i),
\end{equation}
 and 
\begin{equation}
    loss_{concat} = -1/N \sum_{i=1}^N gt^i log(y_{concat}^i),
\end{equation}
 
\noindent where $N$ represent the number of data points, $gt^i$ the ground-truth, which is 0 or 1 if the patches are the same, and $y_{corr}^i$ and $y_{concat}^i$ are the similarity probabilities.

The total loss function is given by the sum of both losses in both heads by

\begin{equation}
    loss_{total} = loss_{corr} + loss_{concat}.
\end{equation}

\subsection{Disparity estimation}
To evaluate the disparity, a maximum disparity value $d_{max}$ is established. To form a wider thermal patch, with the same height of $36$ and width of $36 + d_{max} $, half of this distance is added to both sides of the center point of the patch. With this, the network is able to perform $d_{max}$ translations of a $36 \times 36$ thermal image patch while the visible patch remains the same.

After passing these patches in the feature extractor, $F_{RGB}$ will be a feature map of size $36 \times 36 \times 64$ and $F_{LWIR}$ will be a feature map  of $(36 +d_{max}) \times 36 \times 64$. Next, the $F_{LWIR}$ are passed through the fusion operations and passed through the fully connected layers, as explained earlier. The resulting $y_{corr}$ and $y_{concat}$ correspond to the probability of the patches being the same or different.

For every possible disparity value in the enlarge thermal patch, there is now a matching probability indicating whether the $36\times 36$ patch at this disparity value corresponds to the visible patch or not. The disparity is then the index $\hat{d}$ with the highest probability. This is given by

\begin{equation}
    \hat{d}_{corr} = argmax(y_{corr}),
\end{equation}

and 

\begin{equation}
    \hat{d}_{concat} = argmax(y_{concat}).
\end{equation}

The final disparity is an average of the best disparity from each branch $\hat{d}_{corr}$ and $\hat{d}_{concat}$ and is given by
\begin{equation}
    \hat{d} = \frac{\hat{d}_{corr} + \hat{d}_{concat}}{2}.
\end{equation}

\section{Experiments}

In this section, we provide a detailed overview of the experimental setup, datasets used for training and testing our model, as well as our results with comparison with other state-of-art methods. We also present an ablation study.

\subsection{Implementation details}

Our network is built using the PyTorch framework, with a default patch size of $36\times 36$ with a maximum disparity of 64 for testing. The HRNet backbone was pre-trained on ImageNet \cite{deng2009imagenet}. We only use the first three stages. 

We employ the Adam optimizer for backpropagation. We use a gradient step of 0.001. We trained for 200 epochs with a batch size of 24, as it is the maximum that fits on an RTX 2080 GPU.   

\subsection{Dataset and metrics}

We used two datasets: the LITIV 2014 \cite{st2019online} dataset and the LITIV 2018 \cite{st2019online} dataset. The limited availability of visible-thermal infrared datasets pose a significant challenge for training CNNs. In our particular case, despite using the LITIV datasets, the number of ground-truth points is only slightly above 40,000, which is inadequate for robust training without data augmentation. Therefore, we use two data augmentation techniques. The first data augmentation technique consists of assigning the same disparity as a ground-truth point to its immediate neighbors \cite{beaupre2021domain}. Therefore, for a pixel  with a Manhattan distance of one, we consider that they all have the same disparity. This makes the dataset 5 times bigger. Another technique used to generate more data is mirroring over the $y$ axis. This additionally doubles the number of data points.

Our method was evaluated with cross-validation and trained/tested using different folds, mixing both datasets for training, validation and testing. We used the same folds as Beaupré et al. \cite{beaupre2021domain}. The datasets feature several actors moving in a room. It is to be noted that a few files are missing from the original datasets. Therefore, for a fair comparison, we re-ran the Beaupré et al. \cite{beaupre2021domain} method on the slightly incomplete dataset. 
We have evaluated our method with the recall metric given by
    \begin{equation}
        Recall = \frac{1}{N}\sum_{i=1}^N |\hat{d}^i - gt^i| \leq n,
        \label{recal_formula}
    \end{equation}
\noindent where $N$ stands for the number of points to be evaluated, $\hat{d}^i$ represents the evaluated disparity at a given point, $gt^i$ is the ground-truth at the same given point, and lastly, $n$ represents the allowed correspondence error in pixels.

\begin{table*}[h]
             \caption{Results on LITIV 2014 compared to SOTA Methods. The results are the mean of the 3 folds \cite{beaupre2021domain} domain with standard deviation. \dag We re-ran their code with the slightly incomplete dataset. \ddag: results on the complete dataset. VisiTherS-scales: feature maps concatenated from three scales. VisiTherS-stages: high-resolution feature maps concatenated from two stages.\textbf{ Bold: Best Result} } 
            \centering
            \begin{tabular}{|c||c|c|c|}
            \hline
            Method                                                                       & $\leq$ $1$ $pixel$ $error$ & $\leq$ $3$ $pixels$ $error$ & $\leq$ $5$ $pixel$ $errors$        \\ \hline
          
            Domain Siamese CNN \cite{beaupre2021domain} \dag            & 56.3 $\pm$ 3.6           & {89.9 $\pm$ 0.4}  & 98.5 $\pm$ 0.4                  \\ \hline
            Siamese CNN \ddag \cite{Beaupre_2019_CVPR_Workshops}                               & -                          & $77.9 \pm 5.0$            & -                                 \\ \hline
            St-Charles \cite{st2019online} \ddag                       & 48.2   $\pm$  4.0                   & -                       & -                                 \\ \hline
            Mutual Information \cite{bilodeau2014thermal} ($40 \times 130$) \ddag                       & -                          & 83.3                       & -                                 \\ \hline
            Mutual Information \cite{bilodeau2014thermal} ($20 \times 130$)\ddag                        & -                          & 77.5                       & -                                 \\ \hline
            Mutual Information \cite{bilodeau2014thermal} ($10 \times 130$) \ddag                       & -                          & 64.9                       & -                                 \\ \hline
            Fast Retina Keypoint \cite{bilodeau2014thermal}($40 \times 130$)   \ddag                    & -                          & 64.1                       & -                                 \\ \hline
            Local Self-Similarity \cite{bilodeau2014thermal, st2019online}($40 \times 130$)\ddag                      & 22.6 $\pm$ 10.7                          & 73.4                       & -                                 \\ \hline
            Sum of Squared Difference \cite{bilodeau2014thermal}($40 \times 130$) \ddag                 & -                          & 65.6                       & -                                \\ \hline
            4D-MultispectralNet \cite{duplessis20224d}                                                                 & {57.5 $\pm$ 2.3}  & 88.7 $\pm$ 1.0           & {98.6 $\pm$ 0.4}         \\ \hhline{|=||=|=|=|}

            VisiTherS-scales (ours)                                                              & 
            \textbf{75.0 $\pm$ 0.7}  & 96.2 $\pm$ 0.4           & 99.6 $\pm$ 0.2        \\ 
            \hline
            VisiTherS-stages (ours)                                                           & 74.1 $\pm$ 1.2  & \textbf{96.9 $\pm$ 0.6}           & \textbf{99.8 $\pm$ 0.1}        \\ \hline
           \end{tabular}
            \label{method_comp_2014}
            \end{table*}

\subsection{Comparison with state-of-the-art methods}
The performance of our proposed VisiTherS approach, which incorporates both scale concatenation (VisiTherS-scales) and stage concatenation (VisiTherS-stages), was evaluated against several state-of-the-art (SOTA) methods on the LITIV 2014 and LITIV 2018 datasets. Tables \ref{method_comp_2014} and \ref{toutes_method_comp_2018} present the results of these evaluations. The tables report the mean of three folds. VisiTherS obtains SOTA results on both datasets, with significantly improved performance for the $\leq$ $1$ $pixel$ $error$ and $\leq$ $3$ $pixel$ $error$, particularly for the LITIV 2014 dataset. Given, the low standard deviation, this performance is observed across all folds. This validates our hypothesis that high-resolution features are important for matching the content of dissimilar modalities, like thermal infrared and visible images. Comparing our proposed two versions of feature exploitation strategies, we can observe that they give results that are quite similar with a small advantage to VisiTherS-scales for the $\leq$ $1$ $pixel$ $error$ on LITIV 2014 and the reverse on LITIV 2018. This suggests that incorporating multiple scales can improve the correspondence process since the complexity of the content of patches may differ across scales, but considering different stages can give equivalent results. On the LITIV 2018 dataset, 4D-MultispectralNet that uses object masks is not far behind VisiTherS for the $\leq$ $1$ $pixel$ $error$, but having high-resolution features proves to be globally a better strategy. Adding masks to VisiTherS did not improve our results. 

It should be noted that the results obtained with the Domain Siamese CNN method \cite{beaupre2021domain} differ slightly from those reported in the corresponding paper, as the code was re-run. It yields slightly lower results for $\leq$ $3$ $pixel$ $error$ precision, but for $\leq$ $1$ $pixel$ $error$ and $\leq$ $5$ $pixel$ $error$ precision, the results are higher than their initial study due to differences in the dataset. Considering both the new results and the originals, our proposed method outperforms Domain Siamese CNN significantly showing the benefit of high-resolution features.

\begin{table*}[h]
             \caption{Results of all our methods on the 2018 LITIV dataset. The results are the mean of the three folds with standard deviation.\textbf{ Bold: Best Result.} }
            \centering
            \begin{tabular}{|c||c|c|c|}
            \hline
           Methods                                                                       & $\leq$ $1$ $pixel$ $error$ & $\leq$ $3$ $pixels$ $error$ & $\leq$ $5$ $pixels$ $error$       \\ \hline
            DASC Sliding Window  \cite{st2019online}             & 10.4         & -   & -                  \\ \hline
            Multispectral Cosegmentation  \cite{st2019online}             &26.5         & -   & -                   \\ \hline
             Domain Siamese CNN \cite{beaupre2021domain} \dag            & 44.2           & -  & -                 \\ \hline
               4D-MultispectralNet \cite{duplessis20224d}                                                                  & 60.5 $\pm$ 4.4  & 87.4 $\pm$ 2.0           & 98.7 $\pm$ 0.1         \\ 
               \hhline{|=||=|=|=|}
        
             VisiTherS-scales (ours)                                                                & {63.3 $\pm$ 7.0}  & 92.6 $\pm$ 2.3           & 99.7 $\pm$ 0.2       \\ \hline

           VisiTherS-stages (ours)                                                             & \textbf{63.6 $\pm$ 5.4}  & \textbf{94.8 $\pm$ 2.6}           & \textbf{99.9 $\pm$ 0.1}        \\ \hline
            
            \end{tabular}
            \label{toutes_method_comp_2018}
            \end{table*}\textbf{}

\subsection{Ablation study}

\subsubsection{Ablation study of feature fusion}
Previous studies showed that using concatenation and correlation of features simultaneously gave better results than each separately \cite{beaupre2021domain}. Our new approach was able to validate this observation. In this study, while both convolutional neural networks (CNNs) extracted features from each patch, only one feature fusion operation was performed at a time to observe its performance. This study was performed with VisiTherS-scales. Results are presented in Table \ref{ablation_study_adapt_hrnet}. They indicate that generally, the combination of both fusion methods yields superior performance compared to each fusion operation used separately. However, the correlation fusion method outperformed the concatenation method and the combined method (VisiTherS-scales) for the third fold of LITIV 2014. Nevertheless, by comparing the results for LITIV 2014 in Table \ref{ablation_study_adapt_hrnet}, it can be observed that combining the two operations gives better results than using the concatenation or correlation operation for most folds. For LITIV 2018, the correlation operation outperforms the combined operations for the second fold. The correlation operation performs better in terms of recall metric across all three precision values. In general, correlation is a more efficient approach than concatenation. However, the result is improved when both are used together.

\begin{table*}[h]
            \caption{Ablation study of fusion methods in terms of recall for several pixel errors. \textbf{ Bold: Best Result.} }
            \centering
            \resizebox{\textwidth}{!}{%
            \begin{tabular}{|l||ccc||ccc||ccc|}
            \hline
                   & \multicolumn{3}{c||}{Correlation}                      & \multicolumn{3}{c||}{Concatenation}                    & \multicolumn{3}{c|}{VisiTherS-scales (both operations)}                              \\ \hline
                   & \multicolumn{1}{c|}{$\leq$ $1$ $pixel$ }    & \multicolumn{1}{c|}{$\leq$ $3$ $pixels$ }    & $\leq$ $5$ $pixels$     & \multicolumn{1}{c|}{$\leq$ $1$ $pixel$ }    & \multicolumn{1}{c|}{$\leq$ $3$ $pixels$}    & $\leq$ $5$ $pixels$     & \multicolumn{1}{c|}{$\leq$ $1$ $pixel$ }    & \multicolumn{1}{c|}{$\leq$ $3$ $pixels$}    & $\leq$ $5$ $pixels$     \\ \hline
            LITIV2014-fold1 & \multicolumn{1}{c|}{73.44} & \multicolumn{1}{c|}{\textbf{96.22}} & 99.76 & \multicolumn{1}{c|}{69.22} & \multicolumn{1}{c|}{93.32} & {99.11} & \multicolumn{1}{c|}{\textbf{75.61}} & \multicolumn{1}{c|}{{95.97}} & \textbf{99.80} \\ \hline
            LITIV2014-fold2 & \multicolumn{1}{c|}{{68.23}} & \multicolumn{1}{c|}{{94.89}} & {99.37} & \multicolumn{1}{c|}{64.88} & \multicolumn{1}{c|}{95.45} & 99.41 & \multicolumn{1}{c|}{\textbf{75.08}} & \multicolumn{1}{c|}{\textbf{96.06}} & \textbf{99.67} \\ \hline
            LITIV2014-fold3 & \multicolumn{1}{c|}{77.15} & \multicolumn{1}{c|}{95.87} & 99.66 & \multicolumn{1}{c|}{\textbf{80.93}} & \multicolumn{1}{c|}{\textbf{97.68}} & \textbf{99.86} & \multicolumn{1}{c|}{{74.30}} & \multicolumn{1}{c|}{{96.06}}  & {99.36}  \\ \hline \hline
            LITIV2018-fold1 & \multicolumn{1}{c|}{{68.39}} & \multicolumn{1}{c|}{{92.79}} & {99.62} & \multicolumn{1}{c|}{60.70} & \multicolumn{1}{c|}{90.72} & 99.50 & \multicolumn{1}{c|}{\textbf{68.64}} & \multicolumn{1}{c|}{\textbf{94.65}} & \textbf{99.92} \\ \hline
            LITIV2018-fold2 & \multicolumn{1}{c|}{\textbf{59.33}} & \multicolumn{1}{c|}{\textbf{92.21}} & \textbf{99.71} & \multicolumn{1}{c|}{56.07} & \multicolumn{1}{c|}{87.14} & 97.73 & \multicolumn{1}{c|}{55.40} & \multicolumn{1}{c|}{90.16} & 99.52 \\ \hline
            LITIV2018-fold3 & \multicolumn{1}{c|}{{64.14}} & \multicolumn{1}{c|}{\textbf{93.06}} & {99.55} & \multicolumn{1}{c|}{59.95} & \multicolumn{1}{c|}{87.20} & 97.80 & \multicolumn{1}{c|}{\textbf{65.98}} & \multicolumn{1}{c|}{92.83} & \textbf{99.64} \\ \hline
            \end{tabular}
            }
            
            \label{ablation_study_adapt_hrnet}
            \end{table*}

  \begin{table*}[ht]
    \centering
    \caption{Ablation study of comparison between proposed versions of feature extractors in terms of recall for several pixel errors.\textbf{ Bold: Best Result.} }
    \label{methode_comp_multi_couche}
    \begin{tabular}{|l|l|c|c|}
        \hline
        Dataset & Error & VisiTherS-scales & VisiTherS-stages \\
        \hline
        LITIV 2014 & $\leq$ $1$ $pixel$  & \textbf{75.00 $\pm$ 0.66} & 74.14 $\pm$ 1.21 \\
        \cline{2-4}
         & $\leq$ $3$ $pixels$  & 96.24 $\pm$ 0.40 & \textbf{96.94 $\pm$ 0.56} \\
        \cline{2-4}
         & $\leq$ $5$ $pixels$  & 99.61 $\pm$ 0.23 & \textbf{99.87 $\pm$ 0.04} \\
        \hline
        LITIV 2018 & $\leq$ $1$ $pixel$ & 63.34 $\pm$ 7.00 & \textbf{63.55 $\pm$ 5.37} \\
        \cline{2-4}
         & $\leq$ $3$ $pixels$ & 92.55 $\pm$ 2.26 & \textbf{94.83 $\pm$ 2.64} \\
        \cline{2-4}
         & $\leq$ $5$ $pixels$ & 99.69 $\pm$ 0.21 & \textbf{99.90 $\pm$ 0.10} \\
        \hline
    \end{tabular}
\end{table*}

\subsubsection{Comparison of the two proposed feature extractors}

We conducted ablation studies on both versions of our proposed feature extractors. By concatenating the full resolution of the last two stages, we achieved better results, as demonstrated in Table \ref{methode_comp_multi_couche}. Comparing the results on the LITIV2014 dataset, we observed an improvement in precision from $96.24$ to $96.94$ for $\leq$ $3$ $pixel$ $errors$. However, for $\leq$ $3$ $pixel$ $errors$, the precision dropped slightly from $75.00$ to $74.14$, which can be considered relatively similar as the standard deviation overlaps. The precision for $\leq$ $5$ $pixel$ $errors$ improved slightly from $99.61$ to $99.87$. Regarding the results on the LITIV 2018 dataset, VisiTherS-stages always gets better results compared to the VisiTherS-scales.

\subsubsection{Impacts of the choice of layers}

We tested the accuracy of the high-resolution layer according to each stage in HRNet. We can see the results in the table \ref{hrnet_selon_profondeur}. In this table, $x1$ represents the first stage full resolution output, $x2$ represents the second stage full resolution output, and $x3$ represents the third stage full resolution output (see figure \ref{fig:b}). We can see that the best results are split between $x2$ and $x3$. Indeed, for the $\leq$ $1$ $pixel$ $error$, the last stage has better performance. However, for $\leq$ $3$ $pixel$ $error$ and $\leq$ $5$ $pixel$ $error$, stage $x2$ is better. This, therefore, justifies our choice to use the output of $x2$ and $x3$ in VisiTherS-stages.

\begin{table}[h]
\caption{Results on LITIV2014-fold1, according to the depth of the high-resolution stage. \textbf{ Bold: Best Result.}}
\centering
\begin{tabular}{|l|c|c|c|}
\hline
 Error  & \multicolumn{1}{l|}{x1} & x2                         & \multicolumn{1}{l|}{x3} \\ \hline
$\leq$ $1$ $pixel$ & 73.04                   & 76.66                      & \textbf{76.83}                              \\ \hline
$\leq$ $3$ $pixels$ & 95.58                   & \multicolumn{1}{r|}{\textbf{96.64}} & 96.35                              \\ \hline
$\leq$ $5$ $pixels$ & 99.26                   & \multicolumn{1}{r|}{\textbf{99.84}} & 99.76                              \\ \hline
\end{tabular}
\label{hrnet_selon_profondeur}
\end{table}

\section{Conclusion}

This paper introduces a new method for visible-thermal infrared disparity estimation. The proposed model is designed with two versions of feature extractors that employ two streams to extract features independently for each visual and thermal infrared image patch. The first version concatenates features of different scales in one layer, while the second version concatenates high-resolution features of different stages. The model combines the extracted features from both images using two operations, namely correlation and concatenation, to jointly enhance the network performance. Overall, the proposed model, VisiTherS, offers a novel solution for disparity estimation with promising results. Experimental evaluation on public datasets reveals that the proposed method surpasses several SOTA methods.

\section*{Acknowledgments}
We acknowledge the support of the Natural Sciences and Engineering Research Council of Canada (NSERC), [funding reference number RGPIN-2020-04633].

{\small
\bibliographystyle{ieee_fullname}
\bibliography{egbib}
}

\end{document}